\documentclass[conference]{IEEEtran}
\IEEEoverridecommandlockouts
\usepackage{cite}
\usepackage{amsmath,amssymb,amsfonts}
\usepackage{algorithmic}
\usepackage{graphicx}
\usepackage{textcomp}
\usepackage{xcolor}
\usepackage{color}
\usepackage{booktabs}
\usepackage{array}
\usepackage{verbatim}
\usepackage{multirow}
\usepackage{subcaption}
\usepackage{makecell}
\usepackage{tabularx}
\usepackage{url}
\usepackage{orcidlink}

\def\BibTeX{{\rm B\kern-.05em{\sc i\kern-.025em b}\kern-.08em
    T\kern-.1667em\lower.7ex\hbox{E}\kern-.125emX}}

\begin{document}

\title{Large Language Models are \\ Autonomous Cyber Defenders
 \thanks{Proceedings to appear: IEEE CAI 2025 Adaptive CyberDefense Workshop.}}

\author{
    \IEEEauthorblockN{Sebasti\'an R. Castro, Roberto Campbell, Nancy Lau, Octavio Villalobos, Jiaqi Duan, Alvaro A. Cardenas}
    \IEEEauthorblockA{\textit{University of California, Santa Cruz}}
}

\maketitle

\begin{abstract}
Fast and effective incident response is essential to prevent adversarial cyberattacks. Autonomous Cyber Defense (ACD) aims to automate incident response through Artificial Intelligence (AI) agents that plan and execute actions. Most ACD approaches focus on single-agent scenarios and leverage Reinforcement Learning (RL). However, ACD RL-trained agents depend on costly training, and their reasoning is not always explainable or transferable. Large Language Models (LLMs) can address these concerns by providing explainable actions in general security contexts. Researchers have explored LLM agents for ACD but have not evaluated them on multi-agent scenarios or interacting with other ACD agents. In this paper, we show the first study on how LLMs perform in multi-agent ACD environments by proposing a new integration to the CybORG CAGE 4 environment. We examine how ACD teams of LLM and RL agents can interact by proposing a novel communication protocol. Our results highlight the strengths and weaknesses of LLMs and RL and help us identify promising research directions to create, train, and deploy future teams of ACD agents.
\end{abstract}


\section{Introduction}

Cyber threats are evolving rapidly, becoming more complex, frequent, and costly for organizations. A report by Checkpoint shows that attacks are at all-time highs, with average weekly cyberattacks per organization at 1800 compared to 800 in Q3 of 2024 \cite{checkpoint}.

Traditionally, responding to intrusion alerts requires expert human operators who analyze incidents and recover from attacks. As the sophistication of cyberattacks increases \cite{GhostSAM2024}, manually managing security responses becomes increasingly challenging. To address this issue, automation is needed not only to detect but also to respond to attacks in real-time. Autonomous Cyber Defense (ACD) approaches this using AI agents to detect and mitigate attacks. 



ACD agents have been primarily based on Reinforcement Learning (RL). Still, their applications to real-world systems are constrained by (1) limited explainability, (2) limited transferability to other environments (different attackers or different networks), and (3) training challenges due to complex realistic environment creation and RL sample inefficiency~\cite{wolk2022cageinvestigatinggeneralizationlearned}.


To address the limitations of RL-based approaches, we study whether advances in LLMs can disrupt them. First, LLMs can provide explanations for their decisions. Second, they have been trained with data from diverse threat models and networks. Third, we can use pre-trained models (without the need to develop a gym environment).


Before LLM agents can be used for ACD, we need to answer several research questions including; 
How can we train, design, and deploy teams of ACD agents? 
How do LLMs perform as ACD agents compared to RL agents? 
Can LLMs address the limitations of RL for ACD by providing a practical, human-interpretable reasoning of agent strategy? 
Can LLM ACD agents make sound decisions and generalize against different adversaries? 
What challenges must LLMs address towards ACD? 

To answer these questions, we propose the following contributions:

\begin{enumerate}
    \item \textbf{LLM+RL ACD Framework}: We develop the first study on how LLMs perform in multi-agent ACD environments by creating a framework to integrate LLMs into CybORG~\cite{standenCybORGGymDevelopment2021}, the most prominent simulation environment for ACD and RL. We make our framework open source \footnote{\url{https://github.com/r4wd3r/llms-are-acd}}.
    \item \textbf{ACD Multi-agent Communication Protocol}: We introduce and evaluate the first communication protocol for diverse ACD agents (RL and LLM) in our experiments.
    \item \textbf{Evaluation}: We evaluate RL and LLM agents against a diverse adversary set, discuss  LLM advantages over RL to improve ACD agents. We identify promising directions for future research, including model and prompt tuning, and goal-based planning.
\end{enumerate}

\section{Background}

ACD focuses on creating AI agents that can respond to incidents fast and accurately~\cite{kott2023autonomousintelligentcyberdefenseagent,theron2023AicaArch}. Typically, these agents are trained in simulated environments called ACD gyms. These gyms provide a testbed that recreates a simulated network where agents interact with the environment and learn to defend against attackers~\cite{loevenich2024autonomous}. Well-designed gyms offer a variety of scenarios to effectively explore state spaces~\cite{palmer2024deepreinforcementlearningautonomous}, ensure robustness, and simulate realistic scenarios ~\cite{loevenich2024autonomous,lohn2023autonomous}. Examples of gyms include CyberBattleSim \cite{cyberbattlesim},  NASimEmu \cite{NASimEmu}, and FARLAND \cite{molinamarkham2021farland}. However, these frameworks are not all publicly accessible, actively maintained, or are limited to to particular adversarial scenarios.

One of the most relevant ACD environments is CybORG \cite{standenCybORGGymDevelopment2021}, the foundation for the CAGE Challenges \cite{cage4_challenge_announcement}. 
CybORG offers a high-level abstraction for diverse adversary emulation scenarios (e.g., drone and enterprise networks). The CAGE Challenge is an annual Technical Cooperation Program (TTCP) competition to advance the state-of-the-art ACD. Using the CybORG gym, participants in this challenge submit defender agents (blue agents) to remove attackers (red agents) and maximize availability for network users while protecting critical services.

 CAGE challenges evaluate community submissions based on the highest joint reward obtained in the simulations. Previous submissions to CAGE challenges use RL agents.  Its latest edition, the CAGE 4 Challenge, includes a multi-agent environment with limited communication between defensive agents, partial network observability, and a shared reward between agents. In a submission for this challenge, Singh et al.~\cite{singhHierarchicalMultiagentReinforcement2024} proposed a hierarchical RL approach by decomposing the action space into a meta-policy and three sub-policies. Their results show that a strong expert meta-policy can outperform a joint Proximal Policy Optimization (PPO) policy \cite{SchulmanWDRK17PPO}. In another submission, the Cybermonic team proposed KEEP, a variant of PPO with Graph Convolutional Networks (GCN) \cite{cybermonic2025cage4}. KEEP structures observations as heterogeneous graphs of entities (e.g., files, hosts, routers, and ports) where agent actions alter these graphs. At the time of writing, Cybermonic's KEEP is the only open-source solution for the CAGE 4 competition's leaderboard. So, we use it as a baseline for our evaluations.


Recent work has also started exploring LLM agents for ACD. Rigaki et al. \cite{stochasticparrots} propose an LLM-based ReACT attacker for the NetSecGame environment, and Yan et al. \cite{dependmentoring} proposed a closed-source trained RL agent that provides action feedback to an LLM implemented for the CAGE 1 challenge. Yet these approaches have only been implemented in single-agent simulations and have not explored LLM decision making compared with RL. In this paper, we introduce the first integration of both LLM and RL agents on the multi-agent CAGE 4 challenge and propose approaches to analyze LLM reasoning for action selection. We think these steps will take us closer to understanding how generative AI can help us design future ACD agents for realistic incident response.



\subsection{CybORG CAGE 4}
\begin{figure}[!htb]
    \centering
 \includegraphics[width=0.6\columnwidth]{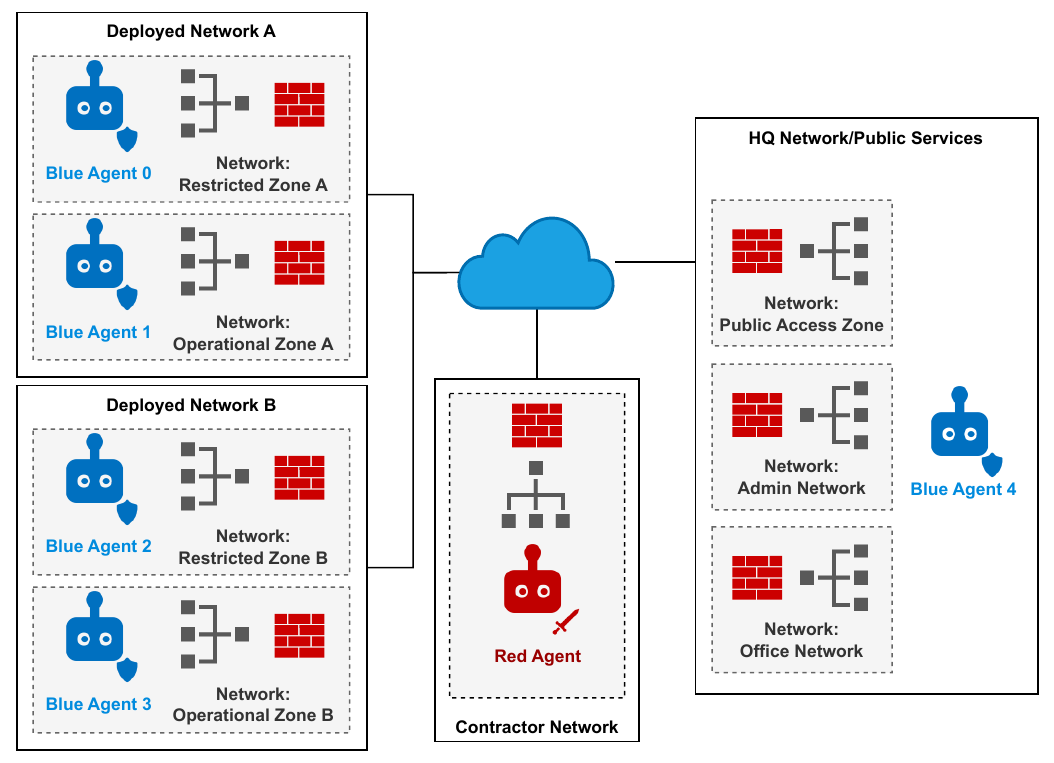}
    
    \caption{CAGE 4 Challenge Architecture }
    \label{fig:cage4_arch}
\end{figure}

The CAGE 4 Challenge simulates a Multi-Agent Reinforcement Learning (MARL) scenario with a team of defender agents (\textit{blue agents}) protecting a network from adversaries (\textit{red agents}) while maintaining service availability for users (\textit{green agents}). As seen in Fig. \ref{fig:cage4_arch}, the network is divided into zones that are protected by the \textit{blue agents} as follows: two deployed networks (A and B), each containing restricted and operational zones; a headquarters network with public access, admin and office zones; and a contractor network where the \textit{red agent} starts. \textit{Green agents} can use all the systems in the network, but they have a slight probability of giving the \textit{red agent} access to a random host by falling victim to a "phishing attack". Each agent has a set of actions defined in Table \ref{tab:actions}.


\begin{table}[!htb]
    \centering
    \scriptsize
    \renewcommand{\arraystretch}{0.9}
    \begin{tabular}{l l l}
        \toprule
        \textbf{Blue} & \textbf{Red} & \textbf{Green} \\
        \midrule
        \textit{Monitor} & \textit{Discover} & \textit{LocalWork} \\
        \textit{Analyse} & \textit{Exploit} & \textit{AccessService} \\
        \textit{DeployDecoy} & \textit{PrivilegeEscalate} & \\
        \textit{Remove} & \textit{DegradeService} & \\
        \textit{Restore} & \textit{Impact} & \\
        \textit{AllowTrafficZone/BlockTrafficZone} & \textit{Withdraw} & \\
        \bottomrule
    \end{tabular}
    \caption{List of actions per agent}
    \label{tab:actions}
\end{table}

CAGE 4 progresses through 3 phases (Planning, Mission~A, Mission B), each with different communication constraints between network zones. Specific operational zones become isolated in active missions, while restricted zones have limited connectivity. Blue agents must adapt their strategies to changing policies, such as blocking unauthorized connections while ensuring that critical services remain available to legitimate users (green agents).

The reward function for \textit{blue agents} is designed to receive penalties if \textit{green agents} cannot use the resources. This can happen due to defensive actions such as blocking a network or restoring a host (which takes the host out of operation for a while) or when \textit{red agents} impact their systems.

\section{LLM Agents for Autonomous Cyber Defense}

\begin{figure*}[!htb]
    \centering
    \resizebox{0.6\textwidth}{!}{%
        \includegraphics{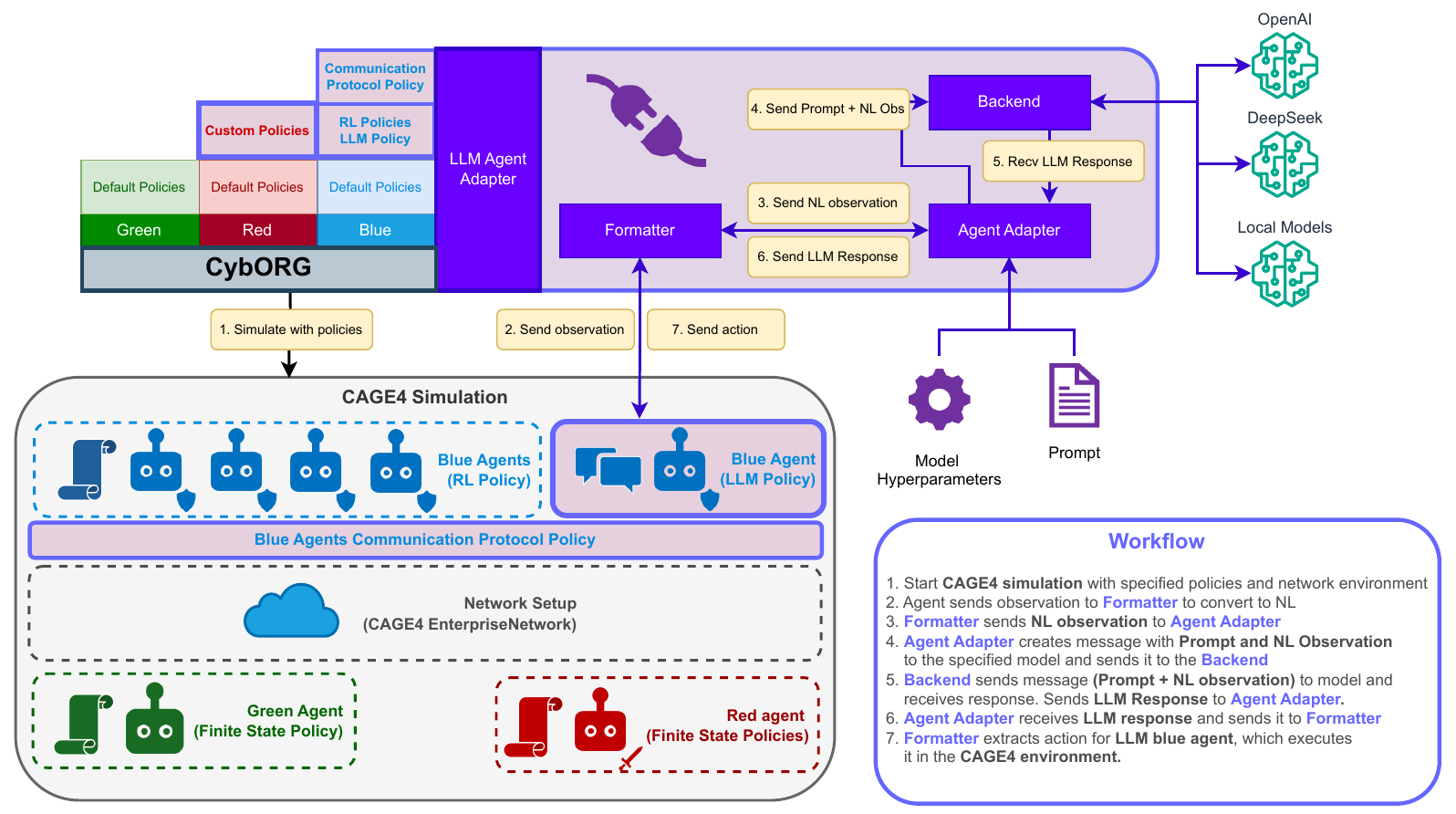}
    }
    \caption{Overview of our LLM Adapter Framework}
    \label{fig:llm_adapater_arch}
\end{figure*}
We created an LLM-based agent for CybORG CAGE 4~\cite{standenCybORGGymDevelopment2021}. To our knowledge, this is the first multi-agent implementation of an LLM interacting with other ACD agent types, including RL ones. In this section, we discuss the challenges and our design. 


\subsection{LLM Adapter Framework}
To understand how LLMs perform as ACD agents in simulated/emulated scenarios, we created an extensible adapter framework to integrate LLMs into CybORG. Its development imposes challenges in terms of performance and cost optimization. During its test phase, CybORG is designed to run with offline post-training RL policies. Thus, the action selection is notably faster than LLMs inference time. In our experiments, LLM inference for action selection with an online small model (i.e., \textit{GPT-4o-mini }\cite{OpenAI}) was in average \textbf{150 times slower} than action selection through a post-training RL policy. To overcome this, when developing initial versions of our adapter, we relied on light offline models (e.g., \textit{TinyLLaMA}) and GPU optimizations, removing costs for paid models and added overhead due to network latency and inference time.


Fig. \ref{fig:llm_adapater_arch} shows our LLM adapter and its workflow for action selection using these models. We highlight our LLM adapter and how it interacts with the agents when the CAGE 4 simulation starts. The figure shows the scenario where one blue agent is LLM-driven and the others are RL-driven. As shown, we add custom policies for both red and blue agents, we include an RL policy based on KEEP GNN for training and testing \cite{cybermonic2025cage4}, we add a novel communication protocol for the blue agents, and propose a flexible LLM adapter to interact with diverse LLMs for action selection.

\subsection{Observation and Response Formatting}
The network state must be formatted in natural language to use an LLM for network defense. Since CybORG is designed for RL, it uses nonhuman-readable observation vectors. To overcome this situation, we built a custom wrapper that adapted CybORG's step observations by parsing them into natural language with the relevant information that each agent sees about their network. 

Table \ref{tab:formatted_obs_llm} shows our observation format per step, which includes the name of the \textbf{Agent}, \textbf{Mission Phase}, the \textbf{Last Action} executed in previous steps and its \textbf{Status} result, the \textbf{Communication Vectors} broadcasted by blue agents, and a list of events under \textbf{Suspicious Activity Detected}.

\begin{table}[!h]
    \scriptsize
    \centering
    \renewcommand{\arraystretch}{1.3}
    \begin{tabular}{@{}l l@{}}
        \toprule
        \textbf{Field} & \textbf{Value} \\
        \midrule
        \textbf{Agent} & \texttt{<agent\_name>} \\
        \textbf{Mission Phase} & \texttt{<value>} \\
        \textbf{Last Action} & \texttt{<action><host/subnet>} \\
        \textbf{Last Action Status} & \texttt{<TRUE/FALSE/UNKNOWN/IN\_PROGRESS>} \\
        \textbf{Communication Vectors} & \texttt{<list:binary\_array>} \\
        \textbf{Suspicious Activity Detected} & \texttt{<list:string>} \\
    \end{tabular}
    \caption{Formatted Observation for LLM agent}
    \label{tab:formatted_obs_llm}
\end{table}

To extract the actions of the blue agent, we format the response of the LLM using a dictionary structure as seen in Fig.~\ref{fig:conversation}. If the LLM responds in an unexpected format that our action parser does not understand, we log an invalid action, and the agent will \textit{Sleep} for that step. This helps us track when the model is not following the response structure we provided in the prompt.


\subsection{Communication Protocol between Defender Agents}

Since blue agents only have visibility in their assigned subnetwork (see Fig.~\ref{fig:cage4_arch}), they need to exchange messages with each other to share threat information. CAGE 4 allows each agent to broadcast a 1-byte vector per step called \textit{Communication Vector}, yet its format is undefined. We use this 8-bit protocol and propose a realistic multi-agent communication strategy.

 Our idea is to summarize the current security level of a network based on each agent's observation and its current state (free or busy). This abstracts the network's general status without substantially increasing the observation dimensions. We propose that each agent should broadcast a message to the others notifying them of (1) suspicious activity detected from another agent's network, (2) the security level of its current network, and (3) its availability for running actions.

For remote activity detection (1), we assign each agent with one single bit since each agent can access only its own observation but can associate the origin of malicious activities from other networks. For the current security level (2), we provide more granularity by assigning 2 bits because the agent's observation contains more dimensions that describe the network security status in more detail. We assign only one bit for availability (3) since the agent can only have free and busy states.

Considering that defender agents are identified from 0 to 4, each blue agent $i$ in the network creates their \textit{Communication Vector} as follows:
\begin{itemize}
    \item \textbf{Bits 0 to 4:} Set $j$ bit to one if malicious action has been detected by $i$ from an agent's $j$ network. Otherwise, zero.
    \item \textbf{Bits 5 and 6:} Level of compromise that agent $i$ identified in its subnetwork: \texttt{00} indicates no compromise, \texttt{01} netscan/remote exploit detected, \texttt{10} user-level compromise in a host, \texttt{11} admin-level compromise in a host.
    \item \textbf{Bit 7:} Set to one if agent $i$ is waiting for an action to finish the execution. Otherwise 0.
\end{itemize}


This protocol can be extended to more detailed summaries depending on capacity and observation features. We believe this approach provides sufficient information to agents to improve their action selection process, especially when agents notify a high-risk security level and other agents detect connections from their network.


\subsection{Prompting}

We aim to evaluate how LLMs act and reason in realistic incident response situations instead of maximizing the CAGE 4 reward function. Therefore, we design prompts that give the LLM a realistic view of a network and avoid mentioning that this as a game, a simulation, or having a reward function. Instead, we provide the agents the role of a "cybersecurity expert defending a network" with formatting rules and examples to guide the reasoning process. \textbf{The detailed prompt is in our repository.}

To evaluate the best prompting strategy that is realistic and minimizes penalties, we use a default CAGE 4 Finite State \textit{red agent}, we set only one \textit{blue agent} as an LLM-driven agent and the rest with a default CAGE 4 defender strategy. We start with an initial \textit{Instructional Prompting}\cite{brownLLMFewShot2020} where we only describe the task and the answer format. Then, we consider the \textit{Few-Shot} strategy where, on top of the task description, we add examples of possible correct answers, which reduces invalid actions. Finally, we explore \textit{Role Prompting}\cite{kongRoleZeroShotReasoning2024}
 by assigning a "cybersecurity expert" role to the agent.

\begin{table}[!htb]
    \centering
    \scriptsize
    \renewcommand{\arraystretch}{0.9} 
    \begin{tabular}{llccc} 
        \toprule
        & \textbf{Model} & \textbf{Role} & \textbf{Fewshot Instruct} & \textbf{Instruct} \\
        \midrule
        & GPT-3.5 Turbo & -4307  & -4349.5  & -4620   \\
        & 4o-mini      & -3810  & -3219    & -2888   \\
        & o1-mini      & -3022  & -3243.5  & -3890.5 \\
        \bottomrule
    \end{tabular}
    \caption{Reward-based prompt tuning with OpenAI Models. Results are average rewards for blue agent. }
    \label{tab:prompt_tuning}
\end{table}

As seen in Table \ref{tab:prompt_tuning}, on average, adding a role to the prompt with few-shot examples may increase the reward with different experiments. Therefore, we decide to use a prompt with a role and few-shot examples.

\begin{figure}[!htb]
    \centering
    \resizebox{0.8\columnwidth}{!}{%
 \includegraphics{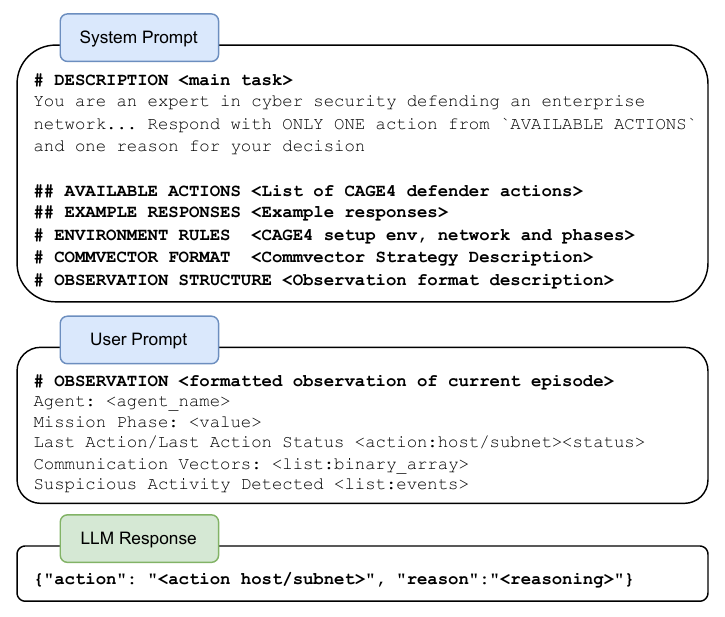}
    }
    \caption{Conversation Workflow for LLM Adapter}
    \label{fig:conversation}
\end{figure}

Figure \ref{fig:conversation} shows an overview of the interaction between the agent and the LLM. We follow the conventions of \textit{system/developer} and \textit{user} messaging for LLMs \cite{OpenAI,DeepSeek,MetaLLaMA}. We set the \textit{System Prompt} to describe the task, response validation, environment rules, and examples. The \textit{User Prompt} contains the formatted observation per step. Note that the \textit{System Prompt} remains consistent during all episodes, whereas the \textit{User Prompt} might change due to the agent's observations.

We fine-tuned the prompt to include clear and concise definitions of each action, including response examples. This is necessary since action names can have multiple meanings, which can mislead the reasoning models. For instance, when we did not define the \textit{Remove} action in our experiments, the LLM assumed it disconnected the host instead of erasing malicious processes.

Our extensible CybORG adapter for LLMs works with the latest online and offline models. For online versions, our current adapter supports OpenAI\cite{OpenAI} and DeepSeek\cite{DeepSeek} through OpenRouter\cite{OpenRouter}. For offline models, it supports open-source models from HuggingFace\cite{HuggingFace}, including Meta's LLaMA\cite{MetaLLaMA} family of models.

\section{Evaluation}
In this section, we describe our experiments and their results. We first describe the methodology and the considerations for the environment setup. Then, we discuss in detail our experiment results and our analysis.

\textbf{Environment and models:} We use the CybORG version included in the CAGE 4 challenge repository \cite{cage4_challenge_announcement} including our custom LLM adapter. We use the latest OpenAI models\cite{OpenAI} \textit{GPT-4o-mini, o1-mini} and \textit{o3-mini}, and the the latest DeepSeek model: \textit{DeepSeek-V3}\cite{DeepSeek}. When running our experiments, we focus on light models to reduce inference time and performance overhead. For all our models, we set the temperature to $1$ to avoid limitations of context dependency from the prompt since it is intentionally not optimized for the CAGE 4 scenario. To keep track of the experiment variables, including reward, standard deviation, and action selection, we use the platform \textit{Wandb (Weights and Biases)}\cite{wandb}. 

\textbf{LLM Messaging:} Since CybORG runs hundreds of steps per episode, and each agent must select an action per step, we use a single \textit{system/user} message to interact with the LLM to reduce token consumption. As seen in Fig. \ref{fig:conversation}, we use one \textit{system} prompt that contains the task to run. We then include the observation of the current step as a \textit{user} message. The LLM-driven agents send a combination of these two messages per step and receive a response in JSON format containing an action and a reason justifying its decision. 

\textbf{Scenarios:} For all our experiments, we run 2 episodes of the game with 500 steps each. Each agent (red, blue, and green) selects an action in each step, and CybORG simulates its execution. We use the default CybORG green agent policy \textit{EnterpriseGreenAgent} while customizing red and blue. In our experiments, we consider four scenarios for blue agents:
\begin{itemize}
    \item \textbf{No blue agents:} All blue agents \textit{Sleep} every turn. Only red and green agents act. This serves as a baseline for our evaluation.
    \item \textbf{All blue agents as LLM:} All blue agents are LLM-driven. They all use the same LLM and the messaging strategy with the same \textit{system} prompt. Their \textit{user} may vary for each agent's observation and include the agent's name to avoid confusing the LLM.
    \item \textbf{All blue agents as RL:} All blue agents are RL-based. We train them by using our own implementation of the communication vector and Cybermonic's KEEP GNN PPO model \cite{cybermonic2025cage4}, which was among the best submissions for the CAGE 4 competition \cite{cage4_challenge_announcement}. This submission did not implement a Communication Vector.
    \item \textbf{1 LLM and 4 RL blue agents:} To balance the RL and LLM distribution, we protect 3 subnets with 1 LLM agent (i.e., \textit{blue\_agent\_4}) and 4 subnets with 4 RL (KEEP) agents. 
\end{itemize}

\textbf{Red agent variants:} CybORG includes limited, finite-state red agents. To reduce the reality gap and diversify the adversary simulation, we implemented the red agent strategies for multi-agent reinforcement learning by Singh et al. \cite{singhHierarchicalMultiagentReinforcement2024}. In particular, we include the following:

\begin{itemize}
    \item \textbf{AggressiveFSMAgent}: Employs an aggressive service discovery action, rapidly scanning the network to identify potential targets. While this approach can quickly map out network services, it increases the likelihood of detection by defensive measures.
    \item \textbf{StealthyFSMAgent}: Utilizes stealthy service discovery actions, aiming to remain undetected while gathering information about the network. This method is slower but reduces the risk of triggering security alerts.
    \item \textbf{ImpactFSMAgent}: Prioritizes action that impacts critical services, focusing on disrupting essential network functions. By targeting high-value assets, the ImpactFSMAgent aims to maximize operational disruption.
    \item \textbf{DegradeServiceFSMAgent}: Prioritizes action that degrades services used by the green agents. This can cause the \textit{GreenAccessService} action to fail. 
\end{itemize}

\subsection{Experiments}
Now we discuss our experiments and analyze our results in terms of the \textit{Performance}, \textit{Reward},  \textit{Reasoning for Action Selection}. Then, we compare the reasoning between an RL and an LLM agent. 

\begin{figure}[!htb]
    \centering
    \includegraphics[width=0.8\columnwidth]{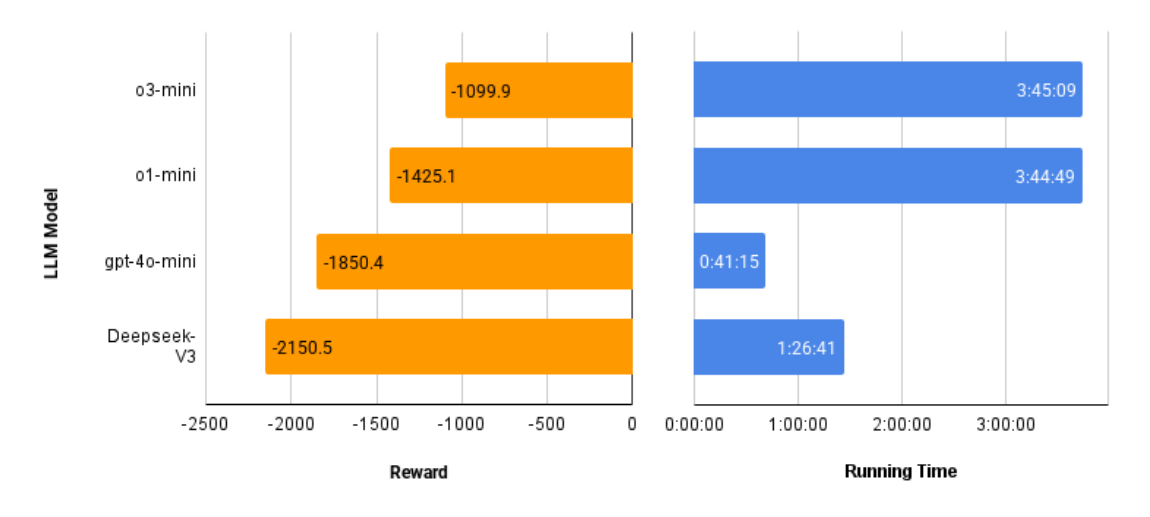}
    \caption{Reward and running-time per model. 1 LLM-driven blue agent and others RL against default \textit{FiniteState} red agent.}
    \label{fig:chart_reward_models}
\end{figure}

\textbf{Performance:}
Regarding general LLM performance, our results show that \textit{GPT-4o-mini} had the more balanced tradeoff between running time and reward. In Fig. \ref{fig:chart_reward_models}, we show the results of our 1LLM+4RL experiment with multiple models. Among the tested models, \textit{o3-mini} achieved the highest reward with the slowest execution time, followed by the bigger \textit{o1-mini} model, which had a similar execution time. \textit{Deepseek-V3} showed the lowest performance in terms of the CAGE 4 reward function but was faster than \textit{o1} and \textit{o3}. Finally, \textit{GPT-4o-mini} had the quickest execution time and but low reward.

Regarding running time, we compared the experiments where all agents were either RL-driven or all were LLM-driven. We did not consider the training phase for the RL-driven agents nor the prompt engineering for task description and observation formatting for LLM-driven ones. On average, our experiments were \textbf{45.2 seconds} long when all blue agents were RL-driven and \textbf{4704.6} seconds when all of them were LLM-driven with the fastest model we tested (i.e., \textit{GPT-4o-mini}). This means that,  without considering the time-consuming training phase for RL agents, RL-driven agents were approximately \textbf{104.1 times} faster than the LLM-driven agents when performing action selection.

\begin{figure}[!htb]
    \centering
    \includegraphics[width=0.8\columnwidth]{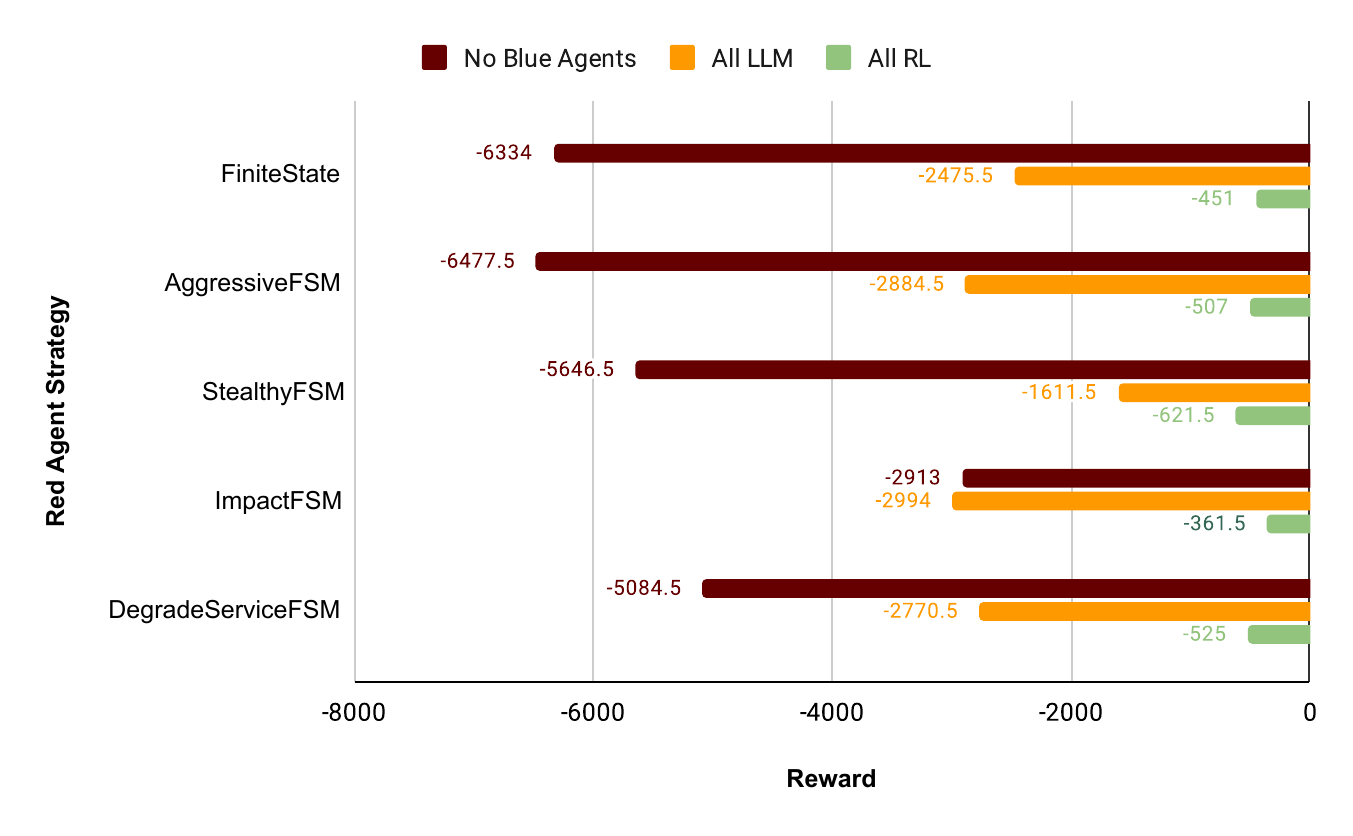}
    \caption{Reward for each red agent strategy against LLM blue agents and RL blue agents.}
    \label{fig:chart_reward_reds}
\end{figure}

\textbf{Reward}: The reward function for CAGE 4 is the cumulative joint reward for \textit{blue agents}, based on penalties for service unavailability. From our experiments with 1LLM+4RL blue agents (Fig. \ref{fig:chart_reward_models}), \textit{o3-mini} received the least penalties. \textit{GPT-4o-mini} had the lowest reward from the OpenAI models, but its execution time was the fastest of all the tested models. Since it is the most balanced, we compare \textit{GPT-4o-mini} against different network agents and the KEEP RL approach.

For CAGE 4, the KEEP RL approach surpassed the all LLM-driven approach with \textit{GPT-4o-mini}. Fig. \ref{fig:chart_reward_reds} shows the reward value with different agents when there are no defenders, when all are LLM-driven (\textit{GPT-4o-mini}), and when all are RL-driven (KEEP). Based on the reward \textit{average \( \mu \)} and \textit{standard deviation \( \sigma \)}, an ALL RL team of defenders performs better against diverse red agents ($\mu = -493$, $\sigma = 95.9$) compared to an ALL LLM team ($\mu = -2547.2$, $\sigma = 498.8$). We see that the diversity of red agents did not substantially affect the reward for RL agents, contrary to the LLM-driven team of agents, which performed marginally better against one adversary (i.e., \textit{StealthyFSM}).

The reward function definition is based on penalties on availability and assumes the attacker is always trying to deny the service to users. However, this is unrealistic since attackers compromise systems without affecting availability \cite{checkpoint}. For instance, with the \textit{ImpactFSM} red agent, the \textit{LLM Agent} performed worse compared to not having any defenders due to the attacker's goal (i.e., prioritize privilege escalation and impact a few compromised hosts). For this scenario, we believe the reward function did not correctly reflect the security level of the network since the adversary had high privileges on the network.

\textbf{Reasoning for Action Selection:} We also want to understand the reasoning behind the LLM agent's action selection. In particular, we compare the RL and LLM agents' action selection under similar scenarios. We first analyze the experiments involving an LLM with the best reward and performance (1 LLM + 4 RL agents, \textit{o3-mini)}. We analyze one episode of 500 steps and remove the initialization step sample.

\begin{figure}[!htb]
    \centering
    \includegraphics[width=0.7\columnwidth]{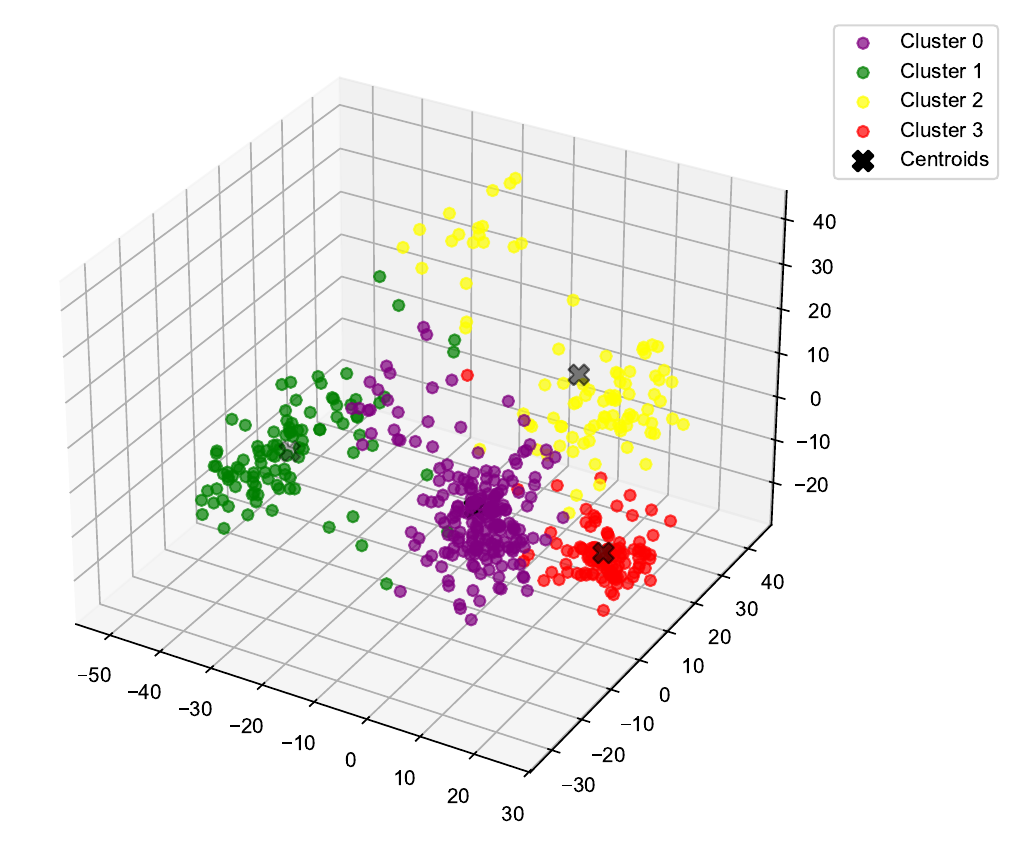}
    \caption{Action Selection Clustering by Reason}
    \label{fig:kmeans_cluster}
\end{figure}

We generate clusters of similar reasons associated with their action to analyze them systematically. We use OpenAI's \textit{text-embedding-3-large} model to convert textual data into numerical embeddings, followed by a PCA-enhanced K-Means\cite{Berkhin2011} clustering strategy. Due to the high dimensionality of our data (499 samples with 3072 features each), DBSCAN\cite{ester1996dbscan} methods were not optimal for our analysis so we used the Elbow Method \cite{elbowMethod} and the Silhouette Score \cite{Rousseeuw1987} and determined $K=4$ as the optimal number of clusters. The application of PCA\cite{Weber1963} helps us reduce the dimensionality to three components, facilitating its visualization and interpretability. Fig. \ref{fig:kmeans_cluster} shows the generated clusters.

\begin{table}[!htb]
    \centering
    \scriptsize
    \renewcommand{\arraystretch}{1.0} 
    \setlength{\tabcolsep}{2pt} 
    \begin{tabularx}{\columnwidth}{lX X} 
        \toprule
        \textbf{Cluster} & \textbf{GPT-4o Summary} & \textbf{Manual Summary} \\
        \midrule
        \makecell[l]{\textbf{Cluster 0 (202 DP)}} 
        & Proactively deploying decoys in various network zones is recommended as a precautionary measure to identify and lure potential red activity early, ensuring detection without impacting service availability amidst ongoing analysis and absence of current suspicious activity. 
        & \textit{DeployDecoy} as a preventative measure and run \textit{Analyse} to detect malicious activity when no alerts are observed. \\ 
        
        \makecell[l]{\textbf{Cluster 1 (103 DP)}}  
        & The main theme of the security-related text cluster is the need for further analysis of repeated INFO-level suspicious connections on various hosts and servers within a network to determine the presence and extent of potential 'red' activity or compromise. 
        & \textit{Analyse} and \textit{DeployDecoy} when perceiving recurrent INFO-level connections; \textit{Remove} as a preventative measure when other blue agents notify malicious activity.\\ 
        
        \makecell[l]{\textbf{Cluster 2 (93 DP)}}  
        & The main theme of the text is that multiple failed decoy deployments in the office network indicate potential configuration issues or undetected threats, necessitating further analysis and reattempts to ensure robust detection and early warning of any red activity. 
        & Attempts to run \textit{DeployDecoy} or \textit{Analyse} have failed. Run \textit{Analyse} to detect configuration issues or malicious activity causing the failure. \\ 
        
        \makecell[l]{\textbf{Cluster 3 (101 DP)}}  
        & The main theme of the text is understanding the importance of continuous monitoring and analysis of decoy deployments in progress to ensure their correct setup and effectiveness in detecting any early red agent activities while maintaining system security and integrity. 
        & \textit{DeployDecoy} action is in progress; then, run \textit{Analyse} to validate the decoy is deployed correctly and reveal early red activity. \\ 
        
        \bottomrule
    \end{tabularx}
    \caption{Comparison of GPT-4o Summaries with Manual Summaries}
    \label{tab:gpt4o_manual_summary}
\end{table}

The clusters and data points (DP) associated with the 499 samples are described in Table \ref{tab:gpt4o_manual_summary}. We structured the action-reason pairs within each cluster into a dictionary format, then used GPT-4o to summarize each cluster and asking the model to "Summarize the main theme of the following security-related cluster in one sentence". We also manually analyzed each cluster to provide a custom description for each.

From a security standpoint, the reasoning for each cluster behind the action selection demonstrates a structured defensive strategy by the LLM-driven agent. Cluster 0 demonstrates a proactive defensive approach by executing analysis and deception actions when no suspicious activity has been detected, and the agent is available. Cluster 1 focuses on actions after receiving alerts and proactively attempts to remove malicious processes if another agent notifies of malicious activity. Cluster 2 shows how the agent tries to understand the failure caused by a previous action by analyzing it. Cluster 3 summarizes the agent's attempts to guarantee the correct deployment of techniques.

\begin{figure}[!htb]
    \centering
    \includegraphics[width=0.9\columnwidth]{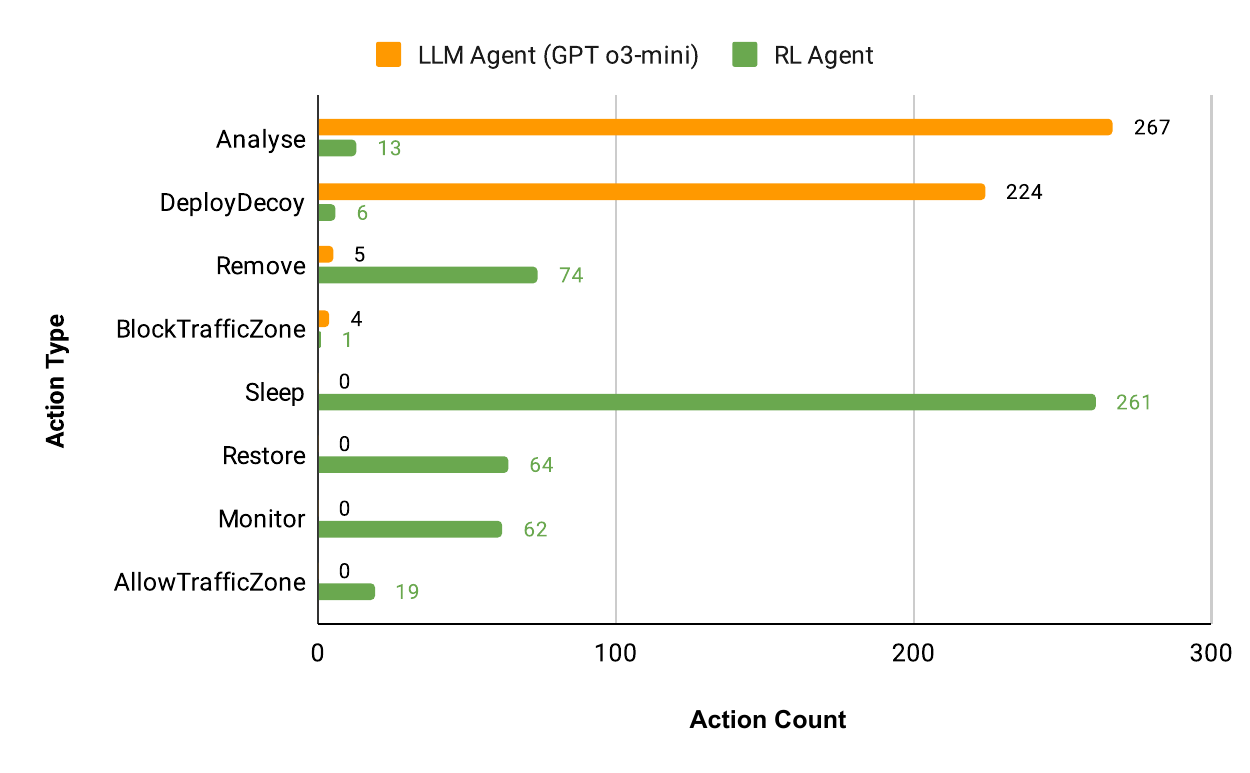}
    \caption{Action Count Comparison for RL and LLM Agents}
    \label{fig:action_count_comparison}
\end{figure}

\textbf{RL and LLM Action Selection Comparison:} We compare the action selection process for an LLM-driven agent with the best reward (\textit{o3-mini}) vs. a KEEP RL-driven agent against the default CAGE 4 attacker. Fig. \ref{fig:action_count_comparison} shows the summary of the action selection for one episode for each agent with a similar scenario.

The KEEP RL agent follows a passive decision-making strategy, relying heavily on \textit{Monitor}, \textit{Sleep}, and \textit{Remove} actions. It frequently transitions into \textit{Sleep}, with a high probability of returning to this state from nearly every other action. When responding to potential threats, the agent prioritizes analyzing hosts for suspicious activity before committing to action.  Actions like \textit{Restore} and \textit{BlockZoneTraffic} appear to be last-resort measures, likely due to their significant impact on availability. By avoiding aggressive interventions unless necessary, the agent ensures smooth network traffic flow and prevents disruptions.

On the other hand, the LLM agent follows a deception and analysis approach, prioritizing the deployment of decoys while avoiding \textit{Restore} actions. After analyzing the reasoning log, we noticed that the LLM-driven agent decided to use \textit{Remove} and \textit{BlockTrafficZone} when other agents notified a possible compromise through their communication vector.  In situations where a user-level compromise was detected elsewhere in the network, the agent prioritizes less disruptive actions like \textit{DeployDecoy} or \textit{Analyse} over \textit{BlockTrafficZone}. The agent avoids the execution of the \textit{Restore} action, possibly because it might cause availability issues for the green users. Our analysis showed that in some cases, the reason for action selection was prone to hallucinations; we saw multiple instances in which the agent misinterpreted the communication vector for an agent (i.e., assumed it was from agent 4 when it was from agent 3) or changed the definition of the action. 

Here, we describe some of the hypotheses we have that justify the different action selection results between RL and LLM agents:

\begin{itemize}
    \item According to the CAGE 4 documentation \cite{cage4_challenge_announcement}, the \textit{Monitor} action is run automatically every turn, and calling it has no effect. Therefore, even though the RL agents selected it, we did not define it as an option in our prompt. 
    \item To maintain the agent active, we did not define the \textit{Sleep} action on the prompt. Including the \textit{Sleep} definition and its use cases when the agent is busy could improve the performance of the LLM agents when actions are in progress.
    \item Our definition of \textit{Restore} and \textit{BlockTraffic} zone may have induced the LLM to avoid them to preserve availability. We described how these actions could disrupt user service. Therefore, adjusting the definitions of actions could diminish the availability of penalties.
    \item To evaluate the inherent LLM reasoning capabilities for ACD, our prompt does not include a strategy with explicit action decision rules nor situation-specific guidance. We could improve the performance of the LLM ACD agent with guidance based on the RL reasoning.
\end{itemize}

As we see, both strategies reflect security reasoning. Despite receiving more penalties than the RL agent, we see that the blue LLM-driven agent could effectively communicate with other RL agents, parse their observations, and reason with a security standpoint without any previous training aside from the provided context in its prompt.

\section{Discussion}

Now, we discuss some limitations of LLMs interacting in multi-agent environments with other ACDs and discuss possible venues to address them.

\textit{Environment Compatibility}: The CAGE 4 environment is designed for RL training. In general, it is not fair to ask a general purpose LLM to select an action without the context an RL-trained approach will have. We are synthesizing the high-dimensional observations of an RL defender for an LLM to reason about it, so an LLM will not observe and interpret the impact of its decisions as well as an RL-driven agent. Moreover, we are conditioning the success of the LLM agent to the reward function designed for the gym. 

\textit{Hallucinations:} Despite describing the simulation rules in our prompt, the LLM agents were prone to hallucinations, including misreading the communication vector observations, compromise levels, and security events. Increasing inference time and fine-tuning models could address this issue with an efficiency drop for action selection. LLM reasoning can improve fast with context-dependent prompting strategies such as Analogical Reasoning\cite{analogical}. Still, it can be prone to hallucinations when the model has not been trained with contextually relevant exemplars. 

\textit{Prompt Definition:} Since our goal is to provide the agent with a realistic context, we design the prompt without specifying the reward policy or providing reward values to the LLM. Adjusting the prompt by describing the reward function used by the simulation and adding context to the simulation constraints (e.g., action execution steps, reward per step/episode, red agents strategies) will increase the expected reward. Still, it will impact the realism of the agent's reasoning for real security scenarios. The prompt can be improved to describe each action in more detail to avoid confusion, as the ones identified with the reason clusters 2 and 3. However, augmenting the length of the prompt will also increase the token consumption.

\section{Conclusions}
We integrate a novel approach to integrate LLM reasoning for defender agents for CybORG's CAGE 4, a realistic ACD multi-agent environment based on RL. We evaluated them against diverse red agent strategies and compared the LLM defending reasoning against pre-trained RL in different scenarios. Our experiments show that LLM ACD agents can communicate effectively when they share a communication protocol. Together, they can protect a network with a security reasoning similar to a team of security operators. 

Despite the discussed LLM limitations, we believe ACD gyms would benefit from LLM reasoning for defenders to bridge their reality gap for transferability and explainability. LLMs can also facilitate changing defensive strategies without retraining by prompt tuning. Defenders can benefit from LLM integrations to RL ACD gyms to train realistic ACD agents' security reasoning without the risks of deploying them on real networks. Our LLM implementation for CybORG facilitates the transferability of optimal policies to other ACD and real environments. This could be done through prompt tuning to reproduce optimal pre-trained RL policies.

\section*{Acknowledgments}
This material is based upon work supported in part by the Air Force Office of Scientific Research under award number FA9550-24-1-0015, and by the National Center for Transportation Cybersecurity and Resiliency (TraCR) USDOT Grant \#69A3552344812. For this research, we also collaborated with OpenAI through the Cybergrant program, using their API credits and funding to advance our research in AI-driven cybersecurity solutions.



\bibliographystyle{IEEEtran}
\bibliography{references}


\end{document}